\documentclass[letterpaper, 10 pt, conference]{ieeeconf}  % Comment this line out if you need a4paper

\IEEEoverridecommandlockouts                              % This command is only needed if 
\usepackage{hyperref}
\hypersetup{bookmarksopen,bookmarksnumbered,
pdfpagemode=UseOutlines,
colorlinks=true,
linkcolor=blue,
anchorcolor=blue,
citecolor=blue,
filecolor=blue,
menucolor=blue,
urlcolor=blue
}
\usepackage{upgreek}
\usepackage{amsfonts,amssymb}
\usepackage{bm}
\usepackage{bbm}

\usepackage{amsthm} % for new theorem
\usepackage{thmtools}
\usepackage{mathtools}
\usepackage{xspace}
\usepackage{units}
\usepackage{booktabs} %for table rulers
\usepackage{tabulary}
\usepackage[usenames,dvipsnames,table]{xcolor} %Options expand the named colours
\usepackage{diagbox}
\usepackage[ruled,vlined,linesnumbered]{algorithm2e}  %updated version of algorithm2e.sty is in ./ folder.
\SetKwComment{Comment}{$\triangleright$\ }{}
\usepackage{ifthen,version}
\usepackage{soul}
\usepackage{csquotes}
\usepackage{microtype}      % microtypography
\usepackage{lipsum}
\usepackage{tikz}
\let\labelindent\relax
\usepackage{todonotes}
\usepackage{enumitem}
\usepackage[noadjust]{cite}
\usepackage{wrapfig}
\usepackage{graphicx}
\usepackage{array}
\usepackage{subcaption}
\usepackage{nopageno}

\usepackage[normalem]{ulem}
\newcommand\redsout{\bgroup\markoverwith{\textcolor{red}{\rule[0.5ex]{2pt}{0.7pt}}}\ULon}

\newcommand{\xxnote}[3]{}
\ifx\hidenotes\undefined
  \usepackage{color}
  \renewcommand{\xxnote}[3]{\color{#2}{#1: #3}}
\fi

\usepackage{multirow}
\usepackage{pbox}

\newtheoremstyle{hypstyle}
{3pt} % Space above
{3pt} % Space below
{\itshape} % Body font
{} % Indent amount
{\bfseries} % Theorem head font
{.} % Punctuation after theorem head
{.5em} % Space after theorem head
{} % Theorem head spec (can be left empty, meaning `normal')

\theoremstyle{hypstyle}

\newcommand{\bbm}{\begin{bmatrix}}
\newcommand{\ebm}{\end{bmatrix}}

\graphicspath{{figs/}}

\title{\LARGE \bf
Reshaping Robot Trajectories Using Natural Language Commands:\\
A Study of Multi-Modal Data Alignment Using Transformers 
\vspace{-2mm} 
}

\author{Arthur Bucker$^{1}$, Luis Figueredo$^{1}$, Sami Haddadin$^{1}$, Ashish Kapoor$^{2}$, Shuang Ma$^{2}$, Rogerio Bonatti$^{2}$\\[2mm]
$^{1}$Technische Universit{\"a}t M{\"u}nchen, $^{2}$Microsoft
\vspace{-3mm}
}

\begin{document}

\maketitle
\thispagestyle{plain}
\pagestyle{plain}

% %%%%%%%%%%%%%%%%%%%%%%%%%%%%%%%%%%%%%%%%%%%%%%%%%%%%%%%%%%%%%%%%%%%%%%%%%%%%%%%

% !TEX root = ../root.tex

% \vspace{-2mm}
\begin{abstract}

Natural language is the most intuitive medium for us to interact with other people when expressing commands and instructions.
However, using language is seldom an easy task when humans need to express their intent towards robots, since most of the current language interfaces require rigid templates with a static set of action targets and commands.
In this work, we provide a flexible language-based interface for human-robot collaboration, which allows a user to reshape existing trajectories for an autonomous agent.
We take advantage of recent advancements in the field of large language models (BERT and CLIP) to encode the user command, and then combine these features with trajectory information using multi-modal attention transformers.
We train the model using imitation learning over a dataset containing robot trajectories modified by language commands, and treat the trajectory generation process as a sequence prediction problem, analogously to how language generation architectures operate.
We evaluate the system in multiple simulated trajectory scenarios, and show a significant performance increase of our model over baseline approaches.
In addition, our real-world experiments with a robot arm show that users significantly prefer our natural language interface over traditional methods such as kinesthetic teaching or cost-function programming.
Our study shows how the field of robotics can take advantage of large pre-trained language models towards creating more intuitive interfaces between robots and machines.
Project webpage: \url{https://arthurfenderbucker.github.io/NL_trajectory_reshaper/}

% \rbnote{a few things to consider for abstract: - Context: What is the context of your work, what is the start of the art 
% - Need: What is the lack of the start of the art 
% - Task: What is the task you want to address in your work (should be 1 sentence) 
% - Object: How do you plan to address your task? 
% - Results: 
% - Conclusions: What are your expected conclusions?}

\end{abstract}

% !TEX root = ../root.tex

%%%%%%%%%%%%%%%%%%%%%%%%%%%%%%%%%%%%%%%%%%%%%%%%%%%%%%%%%%%%%%%%%%%%%%%%%%%%%%%%
% \vspace{-1mm}
\section{Introduction}
% \vspace{-2mm}

Large language models such as BERT~\cite{devlin2018bert}, GPT3~\cite{brown2020language} and Megatron-Turing~\cite{smith2022using} have radically improved the quality of machine-generated text, along with our ability to solve to natural language processing tasks.
Beyond just language, we see a shift in machine learning architectures in multiple domains, as the dominant design paradigm changes from designing task-specific models towards the use of large foundational pre-trained models \cite{bommasani2021opportunities}.
Several of these large models already combine multiple data modalities such as text, images, video, depth, and even the temporal dimension~\cite{radford2021learning,yuan2021florence,alwassel2020self,ma2022compass}.
The use of foundational models is appealing because they are trained on broad datasets over a wide variety of downstream tasks, and therefore provide \textit{general} skills which can be used directly or with minimal fine-tuning to new applications~\cite{bommasani2021opportunities}.

The field of robotics traditionally uses extremely task and hardware-specific models, which have to be re-trained and even re-designed if there are minor changes in robot dynamics, environment and operational objectives.
This inflexible machine learning approach is ripe for innovation with the use of foundational models~\cite{bommasani2021opportunities}, in particular when it comes to task specification in ambiguous scenarios (\textit{What should I do?}) and task learning that can generalize across multiple environments (\textit{How should I do it?}).
Recent works have just started to explored the use of pre-existing foundational models from language and vision towards robotics~\cite{huang2022language,shridhar2022cliport,hong2011recurrent,shao2021concept2robot,goodwin2021semantically}, and also the development of robotics-specific foundational models~\cite{ma2022compass,chen2021learning}.

Our work aims to leverage information contained in existing vision-language foundational models to fill the gap in existing tools for human-robot interaction.
Even though natural language is the richest form of communication between humans, modeling human-robot interactions using language is challenging because we often require vast amounts of data~\cite{fu2019language,hong2011recurrent,stepputtis2020language,goyal2021zero}, or classically, force the user to operate within a rigid set of instructions~\cite{arkin2020multimodal,walter2021language}.
To tackle these challenges, our framework makes use of two key ideas: first, we employ large pre-trained language models to provide rich user intent representations, and second, we align geometrical trajectory data with natural language jointly with the use of a multi-modal attention mechanism.

\begin{figure}[t]
    \centering
    \includegraphics[width=.49\textwidth]{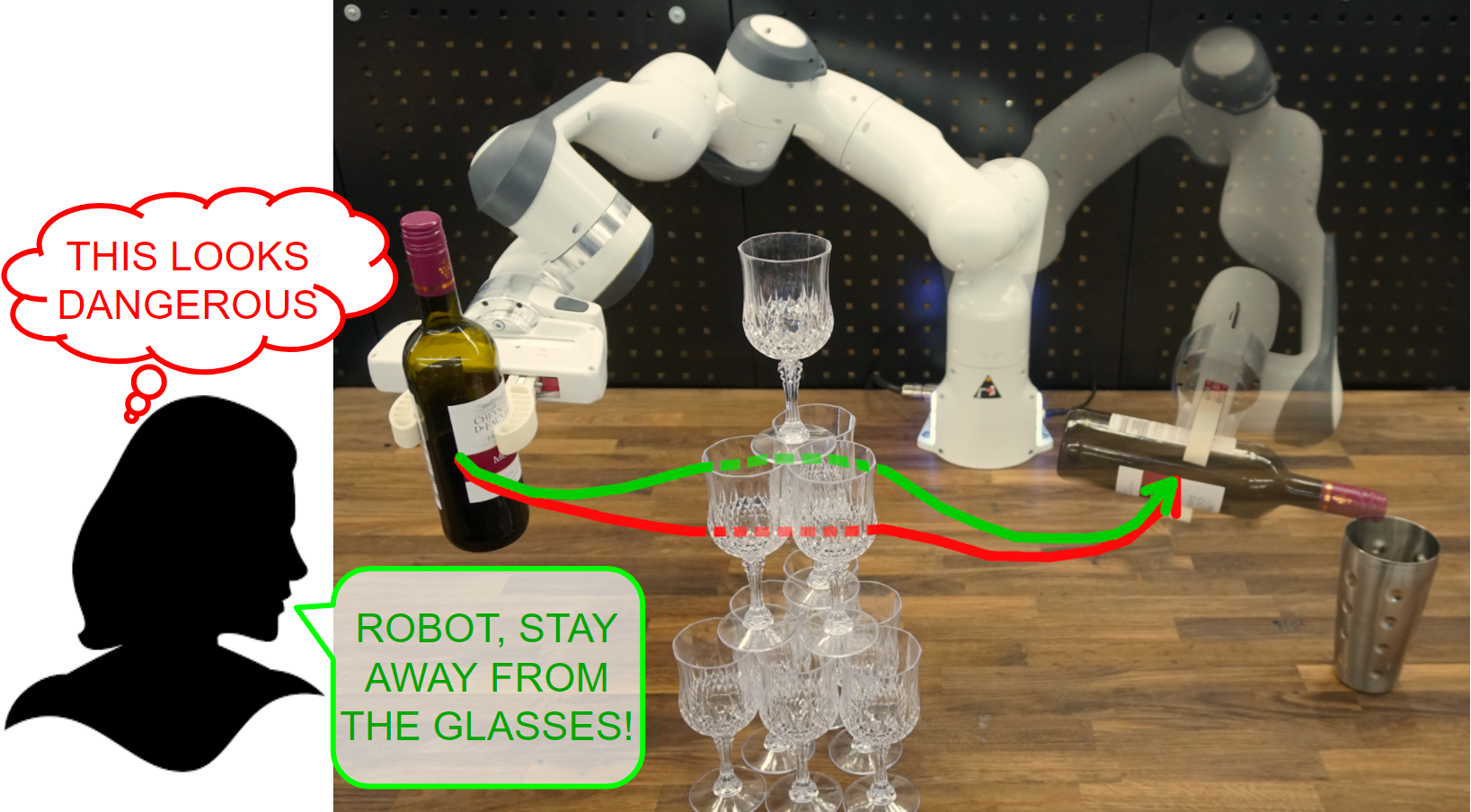}
    \caption{\small{Our system allows a user to send natural language commands to reshape robot trajectories relative to objects in the environment.}}
    \label{fig:main}
    \vspace{-4mm}
\end{figure}

As seen in Fig~\ref{fig:main}, we focus our study on robotics applications where a user needs to reshape an existing robot trajectory according to specific operational constraints.
This class of use cases arises often in human-robot interaction when autonomous agents that employ traditional motion planners (\textit{e.g.} A\textsuperscript{*}, RRT\textsuperscript{*} \cite{Lav06} or MPC~\cite{garcia1989model} concerned solely about obstacle avoidance and dynamics) need to be corrected by a user according to additional semantic or safety objectives. 
For instance, our goal is to enable a factory worker to quickly reconfigure a robot arm trajectory further away from fragile objects, or to allow a user to intuitively tell a robot barista to get a little closer to the cup in order to pour a wine bottle.

% Then main contribution of this paper is to propose a novel transformer-based model architecture for semantic trajectory generation that aligns language features together with geometry information into a predictive trajectory decoder.
Then main contribution of this paper is to propose a novel system with a multimodal attention mechanism for semantic trajectory generation. It can effectively align natural language features with geometrical cues jointly, and perform the goal of trajectory reshaping with a predictive trajectory decoder.
The use of large pre-trained language models to obtain word embeddings allows us to offer a flexible and intuitive user interface, while lowering the requirements on the number of training examples.
% We treat the trajectory generation problem like a sentence generation problem, and not a full parameter regression.
We validate the proposed models in a series of experiments in simulation and in real-world tests with a robotic arm. 
Finally, we show that the proposed trajectory reshaping method is highly preferred by users in comparison with baseline methods both in terms of ease of use and performance.
% \sm{may add one or two more sentences to show the important experimental results.}

% We employ a simulator to generate examples of trajectories that are reshaped using natural language inputs.
% We then use imitation learning to find an alignment between the trajectory data and language. 
% We take inspiration from previous works that created auto-regressive models for X and Y

% \rbnote{However robotics still suffers from rigid structures and models. 
% Give examples from different types of works
% Our goal is to use intuitive human behavior / commands to coordinate a robot?How to shape a robot behavior, during an action execution, through multimodal sparse interactions?
% }

% \rbnote{
% Motivate our specific application. We're reshaping trajectories that were pre-generated by a motion planner. Can be as simple as straight line connecting start and goal, or a trajectory from traditional motion planner but that gets does not obey semantic classes (bartender, manufacturing robot)
% }
% !TEX root = ../root.tex
\section{Related Work} 
\label{sec:related_work}

\textbf{Robots and language:}
As robots become more prevalent in environments outside of laboratories and dedicated manufacturing spaces, it is important to offer non-expert users simple ways of communication with machines.
Natural language is an ideal candidate, given that interfaces such as mouse-and-keyboard, touchscreens and programming languages are powerful, but require extensive training for proper usage~\cite{tellex2020robots}.
Multiple facets of language-based human-robot interaction have been studied in literature, such as
instruction understanding~\cite{macmahon2006walk,kirk2014controlled},
motion plan generation~\cite{stepputtis2020language,lynch2020language,shao2021concept2robot,huang2022language},
human–robot cooperation~\cite{raman2013sorry},
semantic belief propagation ~\cite{arkin2020multimodal,walter2021language},
and visual language navigation~\cite{hong2011recurrent,majumdar2020improving}.
Most of the recent works in the field have shifted from representing language in terms of classical grammatical structure towards data-driven techniques, due higher flexibility in knowledge representations~\cite{tellex2020robots}.

% \rbnote{
% Natural-language-facilitated human–robot cooperation (NLC)
% natural language instruction understanding
% natural language-based execution plan generation
% visual language navigation (batra)
% knowledge-world mapping (action/ context labeling)
% Goal inference vs trajectory shaping
% }

\textbf{Multi-modal robotics representations:}
Representation learning is a rapidly growing field. 
The existing visual-language representation approaches primarily rely on BERT-style~\cite{devlin2018bert} training objectives to model the cross-modal alignments. 
Common downstream tasks consist of visual question-answering, grounding, retrieval and captioning etc.~\cite{sun2019videobert, lu2019vilbert, zhou2020unified, su2019vlbert}. 
% However, these works focus on learning representations only for visual-language domains. 
Learning representations for robotics tasks poses additional challenges, as perception data is conditioned on the motion policy and model dynamics~\cite{bommasani2021opportunities}.
% In robotics area, such visual-language modeling methods have recently been utilized for learning to navigate based on visual-language cues, i.e. visual-language navigation (VLN). 
Visual-language navigation of embodied agents is well-established field with clear benchmarks and simulators~\cite{szot2021habitat,Anderson2018room2room}, and multiple works explore the alignment of vision and language data by combining pre-trained models with fine-tuning~\cite{hao2020genericVLN,thomason2020vision,nguyen2019helpAnna}
 % a task that has been  PREVALENT~\cite{hao2020genericVLN} is proposed as a pretraining and finetuning paradigm for visual-language navigation problems. 
% By learning to align language instructions and visual states for joint representations, three tasks in VLN scenarios can be effectively achieved, i.e. Room-to-room (R2R)~\cite{Anderson2018room2room}, cooperative visual-and-dialog navigation(CVDN)~\cite{thomason2020vision}, and `Help, Anna!' (HANNA)~\cite{nguyen2019helpAnna}. 
To better model the visual-language alignment, \cite{ma2019progressestimation} also proposed a co-grounding attention mechanism.
% was proposed in~\cite{ma2019self}, which is further improved with a progress monitor in~. 
% While these works are all focusing on deploying to specific environments, i.e. indoor home environments, where exists dense object-centric semantics, and explicit relationship between natural language commands and visual scenes can be naturally modeled. 
In the manipulation domain we also find the work of \cite{shridhar2022cliport}, which uses CLIP~\cite{radford2021learning} embeddings to combine semantic and spatial information.
In this paper we also need to align the semantic information with geometry understanding in order to reshape trajectories according to the desired task specifications.
% also focusing on modeling alignments among the triplets of natural language, visual cues and trajectories. 
% The application environments are also not specific to indoor home. In our case, the mapping between each item of the triplets are implicit and usually ambiguous. The natural language instructions can be loosely aligns with the desired trajectory, which makes it more challenging to model such alginments.
% 

\textbf{Transformers in robotics:}
Transformers were originally introduced in the language processing domain~\cite{vaswani2017attention}, but quickly proved to be useful in modeling long-range data dependencies other domains.
Within robotics we see the first transformers architectures being used for trajectory forecasting~\cite{giuliari2021transformer} and reinforcement learning~\cite{chen2021decision,janner2021offline}.
Our work is the first to present a multi-modal transformer model to align visual-language understanding with robot actions for trajectory reshaping.
% , which we call ``\texttt{Trajectory Reshaping Transformer (TR-T)}''.

% \rbnote{
% Visual-language rep
% Trajectory shaping
% Transformers + NLP (large-scale languagne models)
% Transformer + Robotics

% Multimodality
% Human NL instruction are multimodal (Speech, gestures, gaze, physical interactions)  and context dependent 
% }

% By predicting trajectory variations, our work focuses on a combination of task specification and task learning ...
% !TEX root = ../root.tex

\section{Approach}
\label{sec:approach}

\subsection{Problem Definition}
\begin{figure}[b]
    \centering
    \includegraphics[width=.3\textwidth]{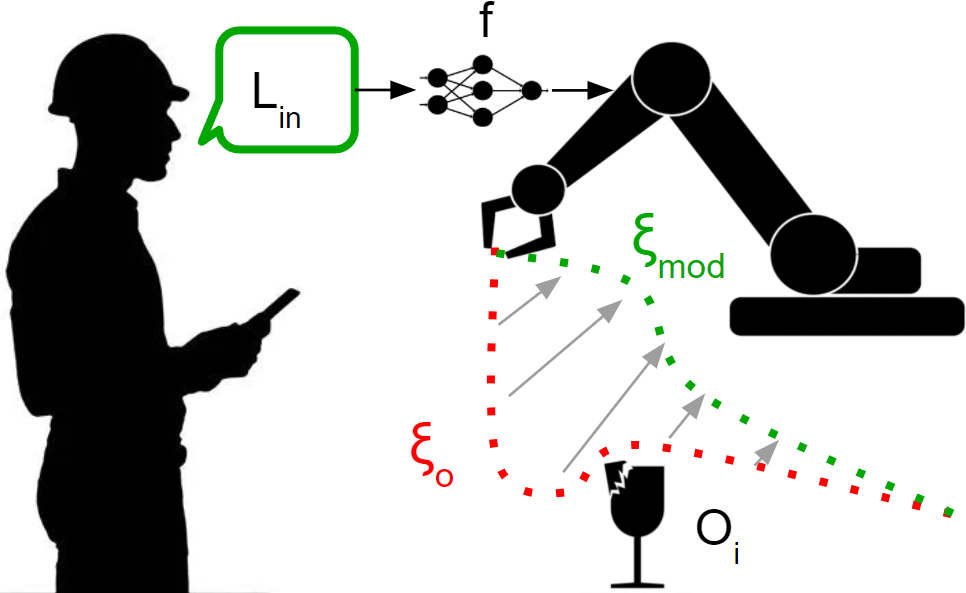}
    \caption{\small{Typical use case for trajectory reshaping. The user's natural language command $L_{\text{in}}$ is processed by function $f$ to reshape the original robot trajectory relative to the target object $O_i$.}}
    \label{fig:example}
\end{figure}

Our overall goal is to provide a flexible language-based interface for human-robot interaction within the context of trajectory reshaping.
One typical application for our systems is that of a user re-configuring a robotic arm trajectory that, although already avoids collisions, gets uncomfortably close to a particular fragile obstacles in the environment.
% We formulate the the trajectory generation problem with a sequential waypoint prediction decoder, which takes into account multiple data modalities from geometry and language into a transformer network.
We design the trajectory generation system with a sequential waypoint prediction decoder, which takes into account multiple data modalities from geometry and language into a transformer network.
The modified trajectory should be as close as possible to the original one throughout its length and respect the original start and goal constraints, while obeying the user's semantic intent.
Fig.\ref{fig:example} depicts the expected model behavior in a typical use-case scenario.

% In this work, we aim at providing a flexible language-based interface for natural human-robot interaction. Specifically, given a natural language commands, e.g. ``stay further away from the glasses'', and an initial trajectory which specifies the path \sm{better in-domain word?}, the model is asked to generate a new trajectory which: 1) executed according to the language command, 2) obey mostly with the initial trajectory, i.e. does not alter the starting point and goal, and roughly consistent with the initial path but just be modified towards the language command. \sm{more accurate definition?} Here, we frame such a task as ``trajectory reshaping'', as shown an intuitive example in Fig.\ref{fig:example} \sm{will need some data samples to draw this figure}.

Let $\xi_{o}: [0,1] \rightarrow \mathbb{R}^2$ be the original robot trajectory, which is composed by a collection of $N$ waypoints $\xi_{o} = \{(x_1,y_1),...,(x_N,y_N)\}$.
We assume that the original trajectory is a reasonable path from the start to the goal positions (\textit{i.e.} avoids collisions) and can be pre-calculated using any desired motion planning algorithm, but falls short of the full task specifications.
Let $L_{\text{in}}$ be the user's natural language input sent to correct the original trajectory, such as $L_{\text{in}}=\text{``Stay away from the wine glass''}$.
Let $\mathcal{O}=\{O_1, ..., O_M\}$ be a collection of $M$ objects in the environment, each with a corresponding position $P(O_i) \in \mathbb{R}^2$ and semantic label, such as $L(O_i)=\text{``glass''}$. Our goal is to learn a function $f$ that maps the original trajectory, user command and obstacles towards a modified trajectory $\xi_{mod}$, which obeys the user's semantic objectives:
\vspace{-3mm}

\begin{equation}
	\label{eq:main}
    \xi_{mod} = f(\xi_{o}, L_{\text{in}}, \mathcal{O})
\end{equation}

\begin{figure*}[t]
    \centering
    \includegraphics[width=.75\textwidth]{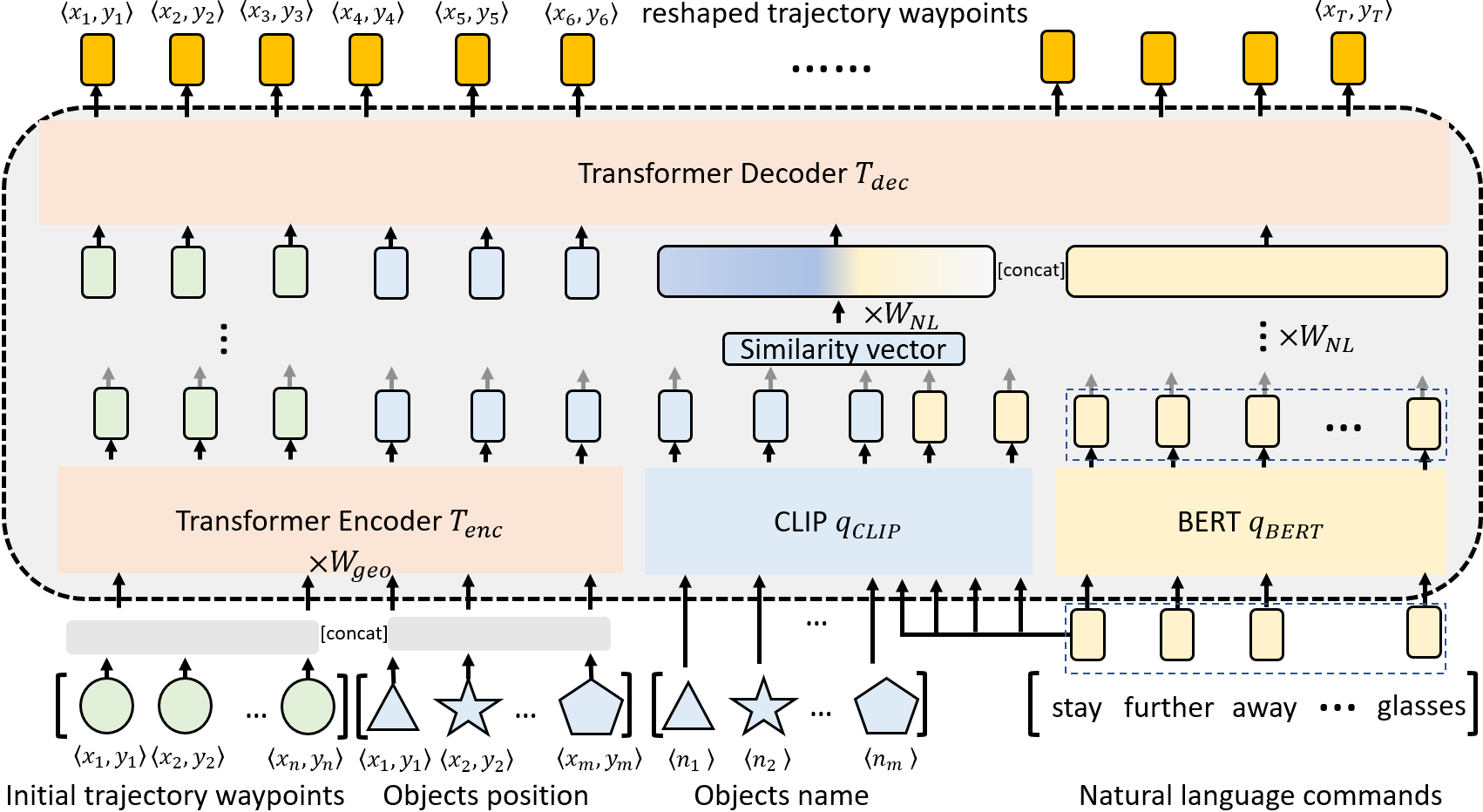}
    \caption{
    \small{We proposed a trajectory reshaping system with multimodal attention mechanism. The whole pipeline consists of pretrianed CLIP and BERT and trainable transformer encoder and decoders. Given the input of initial trajectory (green), objects with their position and semantic labels (blue), and natural language commands (yellow), the model is trained to generate new trajectories in an predictive manner.}
    }
    \label{fig:pipeline}
    \vspace{-1mm}
\end{figure*}

\subsection{Proposed Network Architecture}

We approximate function $f$ from Eq.~\ref{eq:main} by a parametrized model $f_{\theta}$, learned directly from data. 
This mapping is non-trivial since it combines data from multiple distinct modalities, and also ambiguous since there exist multiple solutions that satisfy the user's objective.
Fig.~\ref{fig:pipeline} displays our model architecture, which consists of distinct feature encoders ($q_\text{BERT}$, $q_\text{CLIP}$, $\text{T}_\text{enc}$), whose outputs which are fed into a multi-modal decoder transformer $\text{T}_\text{dec}$ for the sequential prediction of the output trajectory $\xi_{mod}$. In more detail:

\textbf{Language encoding:} 
We use a pre-trained language model encoder, BERT~\cite{devlin2018bert}, to produce semantic features $q_\text{BERT}(z^{\text{in}} | L_{\text{in}})$ from the user's input.
The use of a large language model creates more flexibility in the natural language input, allowing the use of synonyms (shown in Section~\ref{subsec:sim_exp}) and less training data, given that the encoder has already been trained with a massive text corpus.
In addition, we use the pre-trained text encoder from CLIP~\cite{radford2021learning} to extract latent embeddings from both the user's text and the $M$ object semantic labels ($q_\text{CLIP}(z | L)$), which enable us compute a similarity vector between the embeddings, and use this information to identify user's target object. 
In Section~\ref{sec:discussion} we discuss how the CLIP model can potentially be used directly with visual data as opposed to textual object labels.

\textbf{Geometry encoding:}
The original trajectory $\xi_{o}$ is composed of low-dimensional tokens $(x_i,y_i) \in \mathbb{R}^2$.
In order to extract more meaningful information from each waypoint, we apply a linear transform with learnable weights $W_\text{geo}$ that projects each waypoint into a higher dimensional features space, following the example of \cite{giuliari2021transformer}.
The poses $P(O_i)$ of each object are also processed with the same linear transform.
We then concatenate both feature vectors and use a transformer-based feature encoder $\text{T}_\text{enc}$. 
The use of a transformer is preferred for sequences because its architecture can attend to multiple time steps simultaneously, as opposed to recurrent networks, which suffer with vanishing gradient issues~\cite{giuliari2021transformer}.
% \rbnote{details of layers?}

\begin{figure*}[t]
    \centering
    \includegraphics[width=0.9\textwidth]{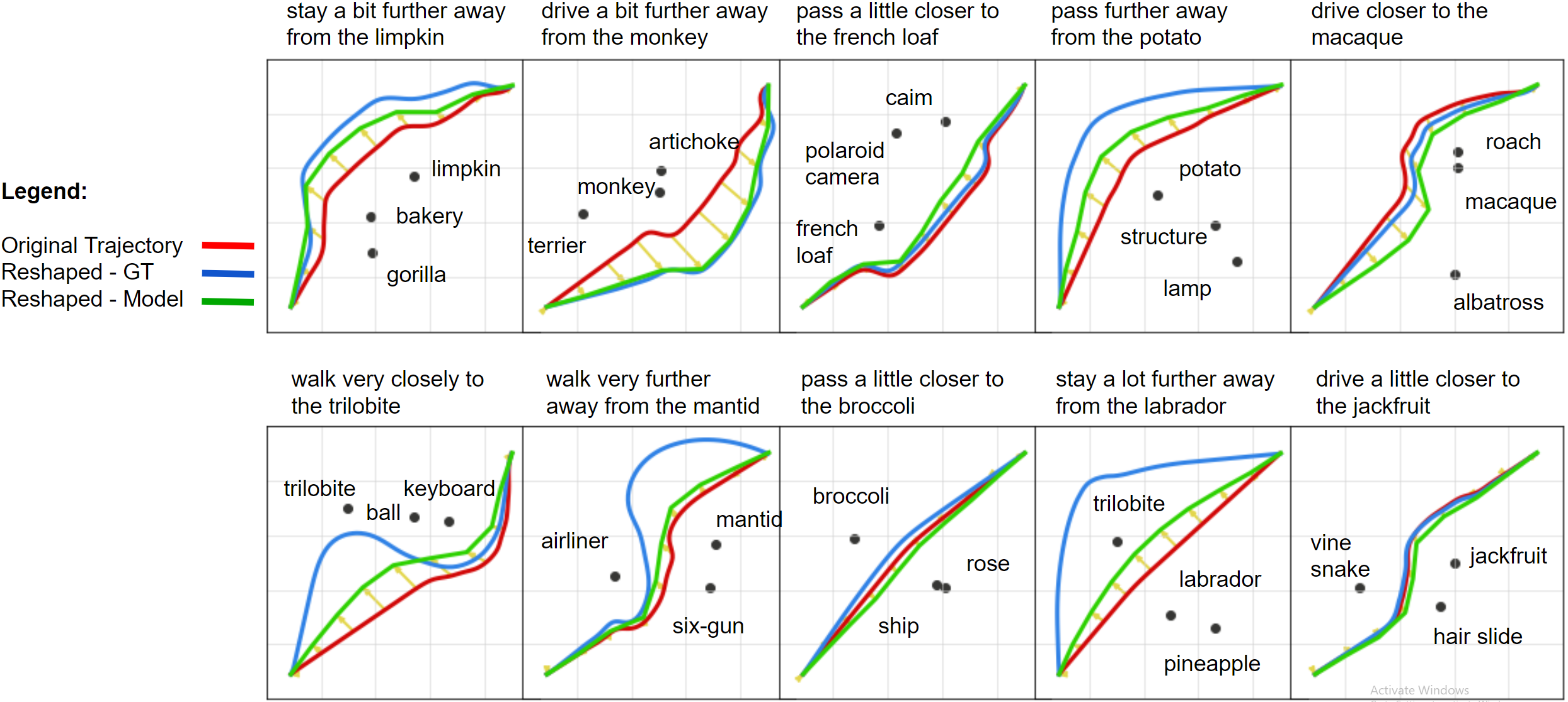}
    \caption{
    \small{Randomly picked planning problems extracted from our validation set. Different colors display the original trajectory (calculated using $A^*$), the ground-truth reshaped trajectory (calculated using CHOMP), and the reshaped trajectory outputted by our model.}
    }
    \label{fig:random_exp}
    \vspace{-2mm}
\end{figure*}

\textbf{Multi-modal transformer decoder:}
Feature embeddings from both language and geometry are combined as input to a multi-modal transformer decoder block $\text{T}_\text{dec}$.
We generate the reshaped trajectory $\xi_{mod}$ sequentially, analogously to common transformer-based approaches in natural language ~\cite{brown2020language,vaswani2017attention}.
Section~\ref{subsec:sim_exp} compares sequential generation with other approaches such regressing to the entire trajectory at once. 
We also verify that a fully-connected architecture cannot achieve the same performance as the transformer-based model.
% \rbnote{details of layers?}
We use imitation learning to train the model, using the Huber loss~\cite{huber1992robust} between the predicted and ground-truth waypoint locations.

\subsection{Synthetic Data Generation}
Data collection in the robotics domain is challenging, specially when we require alignment between multiple modalities such as language and trajectories.
Different strategies range from large-scale online user studies for language labeling~\cite{bonatti2021batteries} all the way to procedural trajectory-language pairs generation using heuristics~\cite{stepputtis2020language}.
Our work relies on a key hypothesis: the use of large-scale language models for feature encoding ($q_\text{BERT}$, $q_\text{CLIP}$) relieves some of the pressure in obtaining a diverse set of vocabulary labels, given that the text encoders are able to find semantic synonyms for different sentence structures.
Therefore, we generated a small but meaningful set of examples with semantically-driven trajectory modifications.
We employed an $A^*$ planner to generate reasonable initial trajectories $\xi_{o}$ in randomized environments with different object configurations, and based on a set of pre-determined semantic combinations, we used the CHOMP motion planner~\cite{ratliff2009chomp} to compute $\xi_{mod}$ by modifying weights of different cost functions. 
Our vocabulary involved different directions relative an object (closer or further away from $\cdot$, to the left/right/front/back of $\cdot$), intensity changes (a bit/little, much, very), and a thousand object labels sampled from the ImageNet vocabulary. We generated a total of $10,000$ trajectory labels.
Fig.~\ref{fig:random_exp} displays examples of original and reshaped trajectories.
% \rbnote{worth putting a figure with a few examples of the ground-truth dataset? or maybe later when we show the training examples?}

\subsection{Implementation Details}

Our language encoders consist of pretrained BERT and CLIP models, with frozen weights.
The output sentence embedding $z^{\text{in}} \in \mathbb{R}^{1\times768}$ is concatenated with the similarity vector of size $1 \times M$ generated from the CLIP embeddings, and scaled to a feature vector of size $1 \times 256$.
For the geometrical inputs we concatenate an array with $M$ object poses with a sequence of $100$ waypoints from $\xi_{o}$, and upscale the features to a vector of size $1 \times 256$.
$\text{T}_\text{enc}$ is a 2-block transformer encoder, and $\text{T}_\text{dec}$ is a 4-block transformer.
Each transformer has 3 hidden layers with $512$ fully-connected neurons in each.
We empirically found it helpful to remove Layer and Batch Normalization~\cite{ba2016layer} steps from both transformers in order to better retain the geometry information.
The model is trained in a two-step fashion: first we augment the training data by randomly rotating and re-scaling the planning problem; second we fix the start and goal locations to always lay in the lower left and upper right corner of the map respectively, and fine-tune the network.
We use the AdamW~\cite{you2019large} optimizer with an initial learning rate $\gamma = 1e-4$ and a linear warm-up period of $15$ epochs. 
We use a Nvidia Tesla V100 GPU with batch size of $64$, and train the model for 500 epochs.
\section{Experiments} 
\label{sec:results}

We execute experiments in both simulated and real environments to evaluate our trajectory reshaping model.
Our goals are the following: 
1) Investigate if the combination of pre-trained large-langauge models together with multi-modal transformers can create efficient and generalizable human-robot interfaces;
2) Quantitatively and qualitatively compare our approach with other classes of human-robot interfaces from the user's perspective; 
3) Demonstrate our natural language method is applicable to real robots.

\subsection{Simulation Experiments}
\label{subsec:sim_exp}

\begin{figure}[th]
    \centering
    \includegraphics[width=.4\textwidth]{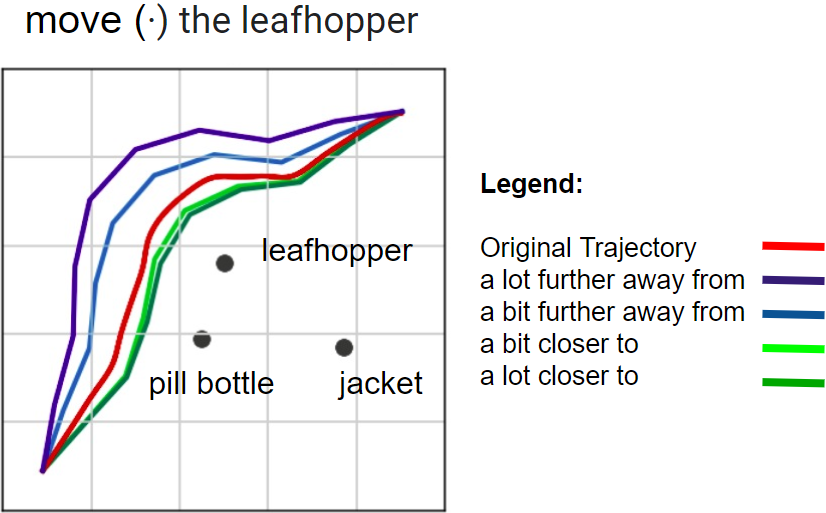}
    \caption{
    \small{Influence of different language inputs to the same environment configuration. The network implicitly constrains changes towards the object, as they would cause unsafe behavior.}
    }
    \label{fig:language_influence}
    \vspace{-2mm}
\end{figure}

We investigate multiple facets of the model's capabilities and performance in a series of simulated experiments:

\textbf{How language influences trajectory behavior:} 
Given a fixed environment configuration, we evaluate the model's ability to follow distinct natural language commands. 
Fig.~\ref{fig:language_influence} displays how a gradient in direction and intensity of language commands correctly modifies the resulting decoded path.

\textbf{How the model behaves in different planning problems:}
To understand the effect of different object configurations and language commands onto the reshaped trajectory, we display in Fig.~\ref{fig:random_exp} a set of randomly sampled planning problems from our validation set. We can see from the image that in most cases the reshaped trajectory correctly models the desired user intent, and falls close to the ground-truth reshaped trajectory.

\begin{figure}[t]
    \centering
    \includegraphics[width=.3\textwidth]{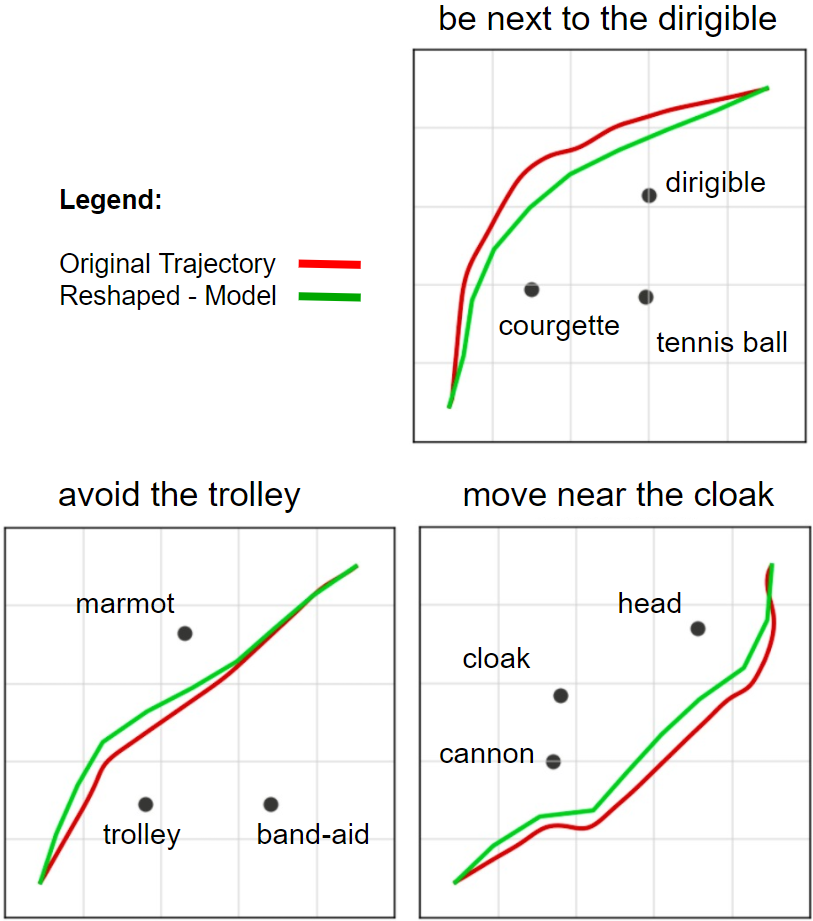}
    \caption{
    \small{Trajectory reshaping results using novel vocabulary (not seen in the training data) as the user input. Our model is able to correctly execute the desired semantic commands due to the large capacity of the BERT and CLIP text encoders.}
    }
    \label{fig:novel_vocab}
    \vspace{-2mm}
\end{figure}

\textbf{Vocabulary and object diversity:}
One key hypothesis assumed true when designing our model architecture was that the use of pre-trained large language models as feature encoders would make our pipeline amenable to a diverse set of natural language inputs, despite the relatively small amount of training examples.
To test this hypothesis we compute results using with novel user commands, with vocabulary not present in our training language labels.
Fig.~\ref{fig:novel_vocab} shows that our model still executes the expected behavior, being able to find the correct semantic meaning despite the new words.

\textbf{Baseline architectures:}
We perform ablation studies comparing different variations of our proposed multi-modal transformer against baseline architectures.
We employ a fully-connected network (FCN) for regression (5 hidden layers, with 512 neurons), that takes as input a single 1D vector composed of the concatenated trajectory, object positions and embedded language features (BERT and similarity vector from CLIP) and outputs a 1D vector with coordinates of all waypoints. The best fully-connected architecture and training procedure was found through a grid-search over the number of hidden layers, neurons, batch size and initial learning rate (64 models in total).
% We executed another grid search to find the best configuration of encoders, decoder and feeding of the embedded textual features for our model. 
Table~\ref{tab:my_label} summarizes the results, and shows that the best architecture is composed by the multi-modal transformer without layer and batch norm, and using a sequential predictive decoder.
The naive predictor referenced in the table shows the loss in the case where we simply copy the original trajectory as the model output, and serves as a baseline loss value. 
Similarly to previous studies in trajectory forecasting~\cite{giuliari2021transformer}, we find that transformers drastically improve model performance, likely due to their unique ability to extract features combining features from distant waypoints.
% One interesting finding, however, was that the use of layer normalization, which is common in transformer architectures, significantly degraded performance. We hypothesize that the use of layer norm loses absolute geometrical information

\begin{table}[h]
    \centering
    \begin{tabular}{l|c|c}
    \toprule
         Model & Features &  Test. Loss \\ \hline
         Naive predictor  &    & 0.0051\\
         FCN w/ regression  &  2.4M  & 0.0025\\
        %  Ours w/o layer norm. &  10.6M &  0.0033 \\
         \textbf{Ours}           &  10.6M & \textbf{0.002}\\
         \bottomrule
    \end{tabular}
    \caption{Baseline architecture comparisons}
    \label{tab:my_label}
\end{table}

% \textbf{What does the multimodal attention layer learn?}
% Fig.~\ref{fig:heatmap} displays a heatmap that gives us insights into what the model is able to learn.
% We can visualize that the attention weights employ features from distinct trajectory locations when evaluating the placement of the next reshaped waypoints.

% \begin{figure}[h]
%     \centering
%     \includegraphics[width=.4\textwidth]{figs/heatmap}
%     \caption{Heatmap of attention weights}
%     \label{fig:heatmap}
%     \vspace{-2mm}
% \end{figure}

\subsection{User Evaluation in Real Robot Experiments}

We also evaluate our system with real-world experiments, and compare our method with the use of multiple human-robot interfaces. 
We use a 7-DOF PANDA Arm robot equipped with a claw gripper, and execute tasks on a $1 \times 1$m tabletop workspace.
A standard desktop computer with an off-the-shelf GPU connected to the robot computes the original trajectories, executes our model, and runs low-level controls for the arm.
We operate the model using 2D planar projections of the original robot trajectories, and respect the original waypoint heights when executing the reshaped motion plans.
We use object positions given by markers, but we discuss the use of vision-based localization in Section~\ref{sec:discussion}.

\begin{figure}[h]
    \centering
    \includegraphics[width=.48\textwidth]{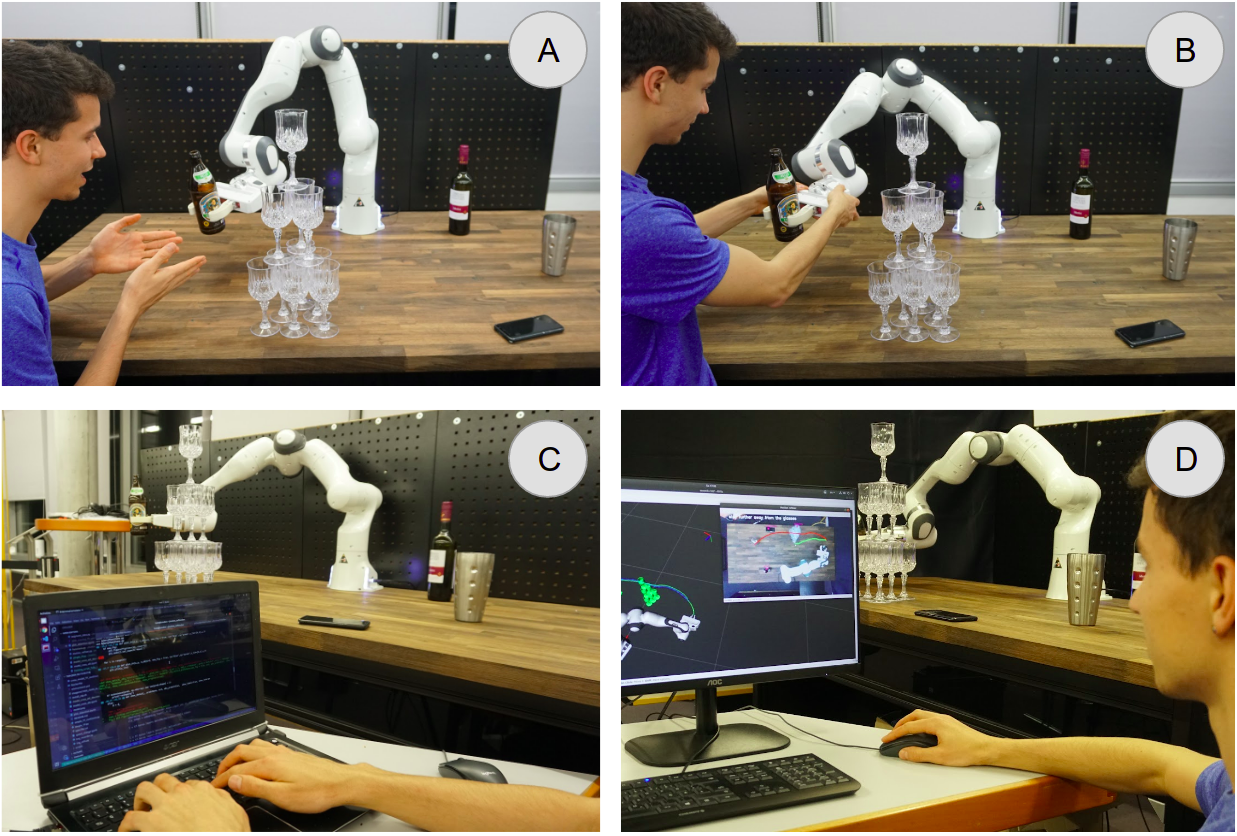}
    \caption{
    \small{Human-robot interfaces tested in the user study: a) natural language (NL), b) kinesthesic teaching (KT), c) programming via keyboard, and d) trajectory drawing.}
    }
    \label{fig:hri_interfaces}
    \vspace{-2mm}
\end{figure}

\begin{figure}[h]
    \centering
    \includegraphics[width=.48\textwidth]{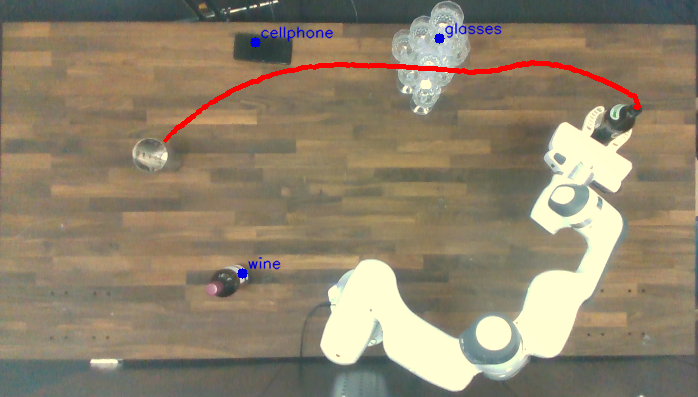}
    \caption{
    \small{Experimental platform used for the user study. The tabletop contains three objects (a cellphone, a wine bottle and crystal glasses), and the original robot trajectory (in red) passes dangerously close to the tower of glasses.}
    }
    \label{fig:platform}
    \vspace{-2mm}
\end{figure}

The goal of the study is to have the user control a robotic bartender. A traditional motion planning algorithm calculates an initial trajectory to transport a bottle of wine towards a cocktail shaker and pour the liquid inside (we leave the problem of learning how to make fancy drinks for future iterations of this work).
This original trajectory comes dangerously close to toppling over a tower of crystal glasses, and the user needs to interact with the robot to make the end-effector trajectory safer.
As seen in Fig.~\ref{fig:hri_interfaces} we test $4$ different human-robot interfaces: natural language (NL-ours), kinesthetic teaching (KT), trajectory drawing (Draw), and programming obstacle avoidance weights via a keyboard and mouse (Prog). A top-down view of the experimental platform is seen in Fig.~\ref{fig:platform}.
All user interactions followed a study protocol approved by the Technical University of Munich's ethics committee, and we conducted a total of $10$ interviews.

\textbf{Quantitative user evaluation:} 
We measured statistics on the number of iterations, success rate and total time taken for users to modify trajectories using the different interfaces.
From Table~\ref{tab:user_study_stats} we see that the programming interface takes by far the longest for users to master, and requires a large number of iterations. 
In the meanwhile, NL is the fastest option.
We see a large number of failures for kinesthetic teaching and drawing because user inputs are often times kinematically infeasible by the robot joints.
The natural language method proveds to be the most robust, and we found no failure cases during in the study.
% \rbnote{then why are number of iter not the smallest?}.
% \abnote{explain the constraint satisfaction module}

\begin{table}[h]
    \centering
    \begin{tabular}{l|c|c|c}
    \toprule
         Interface & Avg. iterations &  Success rate (\%) & Avg. Time (s)\\ \hline
         NL  &  \textbf{1.33}  & \textbf{100} & \textbf{81}\\
         KT   &  1.78  & 56.24 & 139\\
         Draw &  1.89 &  64.7 & 120\\
         Prog  & 4.00  &  91.66 & 284\\ \hline
    \end{tabular}
    \caption{Statistics collected over the user study experiment}
    \label{tab:user_study_stats}
\end{table}

\textbf{Qualitative user evaluation:}
After the experiments we asked users to rate the trajectories produced by different interfaces according to different criteria in a psychometric questionnaire:
\begin{enumerate}  
    \item How satisfied were you with the final robot motion?  
    \item How easy was refining the robot motion?  
    \item How safe was the final robot trajectory?
    \item How natural was the human-machine interaction?
    \item How predictable was the trajectory for you?  
\end{enumerate}
Table~\ref{tab:user_ratings} summarizes the responses. We can see that most methods present a similar user satisfaction level except for programming, which was rated  lower likely due to the difficulty of interaction. NL was rated as the easiest and most natural method, but at the same time was deemed less predictable than KT and drawing because with these two methods users have direct control over the final trajectory.

\begin{table}[h]
    \centering
    \begin{tabular}{l|c|c|c|c|c}
    \toprule
         Interface & Satisfied &  Ease of use & Safety & Natural & Predictable\\ \hline
         NL  &  \textbf{90}  & \textbf{92} & 92 & \textbf{98} & 72\\
         KT  &  \textbf{90}  & 88 & 88 & 78 & \textbf{96}\\
         Draw &  88  & 74 & \textbf{100} & 80 & 88\\
         Prog  &  62  & 58 & 82 & 62 & 48\\ \hline
    \end{tabular}
    \caption{User ratings collected in the user study}
    \label{tab:user_ratings}
\end{table}

\textbf{Final experimental remarks:}
Overall we find from the experiments that our proposed natural language model stands as a strong alternative to traditional human-robot interfaces.
Kinesthetic teaching often not a viable solution for real-world trajectory reshaping depending on the robot's size, form factor and actuator types.
Trajectory drawing is also not a robust solution, as human-defined trajectories often extrapolate acceleration and kinematic constraints.
By combining semantic and geometrical information, our method provides a natural and effective interface for trajectory reshaping.

% If you were to choose between one of the forms of trajectory refinement, which one would you choose? Please order your favorites from 1 (favorite) to 5 (least favorite) in terms of (a) performance of the final trajectory;  (2) easiness to interact with;  (3) time required to refine the trajectory.  

% \begin{itemize}
%     \item \texttt{}
%     \item \texttt{}
%     \item \texttt{Ours}
% \end{itemize}

% After all set the experiments,  for each motion refinement strategy, 
% participants were asked to
% complete a psychometric questionnaire. Participants were asked to grade (1 to 5) the following
% questions (in addition to providing further comments on the experiments):

% \subsubsection{Quantitative evaluation}

% \textbf{Baseline.}

% \subsubsection{Ablation Study}

% \textbf{How the model work on out-of-vocabulary (OOV)?}
% \sm{new vocab}

% \textbf{How the model work on `unseen' objects?}
% \sm{non-exisitng object}

% \subsubsection{User Study}

% !TEX root = ../root.tex

\section{Conclusion and Discussion}
\label{sec:discussion}

In this paper, we present a novel system with multimodal attention mechanism for semantic trajectory reshaping. 
Given flexible natural language commands and an initial trajectory, it can effectively reshape the trajectory consistent with the language commands. 
The multimodal attention architecture provides a way for us to jointly align natural language features and the geometrical cues.

We verify through our experiments that by leveraging large pretrained language models like BERT and CLIP, our proposed system creates a flexible and intuitive user interface.
Given that these foundational models train on massive corpus of data, we are able to train our robotics system with a smaller dataset, and let the language model find similarities between sentences if novel vocabulary is used.

By evaluating our methods on both simulation and real-world application scenarios, we show that our model outperforms baseline trajectory reshaping approaches in terms of loss values and quality of results.
From the user's perspective, we also perform a user study and show that users significantly prefer our natural language interface in opposition to other methods such as kinesthetic teaching or programming interfaces. Our method is faster to use, and results in a higher success rate. 
% approach which can directly take in natural language for reshaping. \sm{other observations.} 

Even though our study does not address the visual modality, we are confident that our current architecture would also be able to align this additional data through the CLIP encoder. For future iterations of this work we're interested in using images of the objects as opposed to directly inputting the semantic labels to the model. 
In addition, in the future we are interested in designing methods that also consider causal relations among objects from the natural language commands in order for the robot to execute more complex task-driven behaviors.
% For future study, to build the model that can directly taken from raw pixel space is preferable, which can enable more practical applications and without the demanding of semantic labels for objects. 

% \rbnote{discuss how images can be used in CLIP}

% \pagebreak
\section*{Acknowledgments}

AB gratefully acknowledges the support from TUM-MIRMI. 
% The authors also thank John Doe for the helpful discussions.

%%%%%%%%%%%%%%%%%%%%%%%%%%%%%%%%%%%%%%%%%%%%%%%%%%%%%%%%%%%%%%%%%%%%%%%%%%%%%%%%

% \bibliographystyle{IEEEtran}
% \bibliography{IEEEabrv,IEEEexample}
\footnotesize{
\bibliographystyle{IEEEtran}
\bibliography{IEEEexample.bib}
}

% \bibliographystyle{IEEEtran}
% {\small
%   \bibliography{IEEEexample}  % .bib
% }

\end{document}